\documentclass[runningheads]{llncs}
\usepackage[T1]{fontenc}
\usepackage{graphicx,verbatim}
\usepackage{amsmath}
\usepackage{booktabs}
\usepackage{multirow}
\usepackage{subcaption}
\usepackage{makecell}
\usepackage{cite}
\usepackage{graphicx}
\usepackage{float}
\usepackage{xcolor}
\usepackage{url}

\begin{document}
\title{CAPRI-CT: Causal Analysis and Predictive Reasoning for Image Quality Optimization in Computed Tomography}
\titlerunning{Causal Analysis and Predictive Reasoning for Image Quality Optimization}
%
\begin{comment}  %% Removed for anonymized MICCAI 2025 submission
\author{First Author\inst{1}\orcidID{0000-1111-2222-3333} \and
Second Author\inst{2,3}\orcidID{1111-2222-3333-4444} \and
Third Author\inst{3}\orcidID{2222--3333-4444-5555}}
%
\authorrunning{F. Author et al.}
\institute{Princeton University, Princeton NJ 08544, USA \and
Springer Heidelberg, Tiergartenstr. 17, 69121 Heidelberg, Germany
\email{lncs@springer.com}\\
\url{http://www.springer.com/gp/computer-science/lncs} \and
ABC Institute, Rupert-Karls-University Heidelberg, Heidelberg, Germany\\
\email{\{abc,lncs\}@uni-heidelberg.de}}

\end{comment}

%\author{}  %% Added for anonymized MICCAI 2025 submission

\author{Sneha George Gnanakalavathy\inst{1}, Hairil Abdul Razak\inst{2}, Robert Meertens\inst{2}, Jonathan E. Fieldsend\inst{1}, Xujiong Ye\inst{1}, Mohammed M. Abdelsamea\inst{1}\thanks{Corresponding author: \texttt{m.abdelsamea@exeter.ac.uk}}}

\authorrunning{Gnanakalavathy, Razak, Meertens, Fieldsend, Ye \& Abdelsamea}
\institute{ Department of Computer Science, University of Exeter,
            UK  \and
            Department of Health and Care Professions, University of Exeter,
            UK  }

\maketitle              % typeset the header of the contribution
\begin{abstract}
In computed tomography (CT), achieving high image quality while minimizing radiation exposure remains a key clinical challenge. This paper presents CAPRI-CT, a novel causal-aware deep learning framework for Causal Analysis and Predictive Reasoning for Image Quality Optimization in CT imaging. CAPRI-CT integrates image data with acquisition metadata (such as tube voltage, tube current, and contrast agent types) to model the underlying causal relationships that influence image quality. An ensemble of Variational Autoencoders (VAEs) is employed to extract meaningful features and generate causal representations from observational data, including CT images and associated imaging parameters. These input features are fused to predict the Signal-to-Noise Ratio (SNR) and support counterfactual inference, enabling “what if” simulations, such as changes in contrast agents (types and concentrations) or scan parameters. CAPRI-CT is trained and validated using an ensemble learning approach, achieving strong predictive performance. By facilitating both prediction and interpretability, CAPRI-CT provides actionable insights that could help radiologists and technicians design more efficient CT protocols without repeated physical scans. The source code and dataset are publicly available at \url{https://github.com/SnehaGeorge22/capri-ct}. %This work advances explainable AI in medical imaging and underscores the potential of causal machine learning for personalized imaging calibration and low-dose optimization.
\vspace{-10pt}

\vspace{-10pt}

% Authors must provide keywords and are not allowed to remove this Keyword section.

\end{abstract}

\vspace{-10pt}
\section{Introduction}
\vspace{-5pt}

Computed tomography (CT) plays a vital role in diagnostic imaging by providing high-resolution cross-sectional images of internal anatomy. However, optimizing image quality while minimizing radiation exposure remains a persistent challenge, particularly in cases involving contrast agents or low-dose protocols. One of the common metrics used to assess CT image quality is the Signal-to-Noise Ratio (SNR), which quantifies the clarity of meaningful signal relative to background noise. Accurate estimation and control of SNR are essential for ensuring diagnostic reliability and patient safety \cite{selfsupervised2019}.

Traditionally, image quality evaluation has relied on empirical testing using physical phantoms and handcrafted calibration rules\cite{goldphantom2021}. While these methods are informative, they are limited in scalability, adaptability, and their ability to account for complex dependencies among imaging parameters such as tube voltage, scan time, and contrast media concentration \cite{mambo2025}. Moreover, conventional deep learning models often lack interpretability and generalizability across diverse imaging setups, mainly due to the loss of cause-and-effect relationships \cite{causalitymatters2020, blackbox2025}. To address these limitations, recent advances in causal machine learning provide powerful tools to uncover underlying causal structures in medical imaging data \cite{review2022, pawlowski2020}. In this work, we introduce a causal-aware deep learning framework that predicts and interprets SNR values in phantom-acquired CT images. %Our key contributions include: 
%\begin{enumerate}
%    \item Integrating image data and metadata using a Causal Variational Autoencoder (CausalVAE), enabling the model to capture the underlying causal structure of imaging conditions.
 %   \item Enabling counterfactual inference, allowing us to simulate \textit{what-if} scenarios that explore how changes in scan parameters or contrast agents would affect image quality \cite{fundus2022}, \cite{latent3d2024}.
  % \item Enhancing model interpretability to improve trust and transparency, thereby facilitating evidence-based optimization of CT protocols \cite{caussl2023}.
%\end{enumerate} 

\vspace{-10pt}
\section{Related Work}
\vspace{-5pt}

Most deep learning models in CT imaging focus on making accurate predictions, such as disease detection or image quality estimation. However, these models often function as black boxes, providing predictions without explaining the underlying causal factors. In healthcare, this lack of interpretability can reduce trust and clinical applicability. To address this challenge, recent research has explored causal machine learning approaches that uncover how different variables influence each other and affect imaging outcomes. Integrating causal inference with deep learning enables not only accurate predictions but also simulation of \textit{what-if} scenarios, which is crucial for designing optimized imaging protocols.

Several recent works have applied causal frameworks in medical imaging. For example, Liao\cite{fundus2022} and Pawlowski \cite{pawlowski2020} used Deep Structural Causal Models to generate counterfactual medical images, improving interpretability in diagnosis by highlighting causal features rather than mere correlations. Similarly, MAMBO-NET \cite{mambo2025} improves segmentation of organs and tumors in CT by focusing on image regions with causal impact, reducing noise from irrelevant background. 

Counterfactual generation has also been explored in brain MRI through latent variable models such as Latent3D \cite{latent3d2024}, providing pathways to disentangle causal factors in complex imaging data. Further, causal diagram approaches have been applied to CT imaging in clinical outcome studies, such as staging chest CT for colon cancer using inverse probability weighting \cite{staging2023}. These examples demonstrate the broad applicability of causal reasoning across imaging modalities and clinical contexts.

Multiple recent studies have also emphasized the importance of addressing dataset bias, domain shifts, and fairness using causal principles in medical imaging \cite{review2022, caussl2023, nofairlunch2023, kaddour2022, highstakes2022, highstakes2022b}. These works highlight how causal learning can mitigate hidden confounders and improve model robustness in safety-critical healthcare applications. In addition, causal frameworks have been explored for multi-interventional learning \cite{multiatt2024}, pancreatic tumor segmentation \cite{deepcausalpancreas2024}, and protocol optimization in CT phantom experiments \cite{goldphantom2021}, demonstrating the growing utility of causal reasoning in CT protocol design.

%Building on this foundation, our proposed CAPRI-CT model uniquely focuses on predicting SNR in CT scans by explicitly modeling the causal influence of scan parameters such as voltage, current, and contrast agent. In addition to CAPRI-CT, we conducted extensive experiments with comparison baseline models including CNN (Convolutional Neural Network), ResNet \cite{resnet2016} and SqueezeNet\cite{squeezenet2016}, each subjected to similar hyperparameter tuning to ensure fairness and comprehensive evaluation of model performance. Our CAPRI-CT model works by understanding how different imaging settings directly affect the SNR value, allowing it to make more accurate and explainable predictions. Baseline models, on the other hand, just find patterns in the data without considering these cause-and-effect relationships, which can make their predictions less reliable.

\vspace{-10pt}
\section{Data acquisition}
\vspace{-5pt}

The dataset comprises CT images of a custom-designed Perspex phantom containing 169 cylindrical holes of varying diameters (4–7 mm) arranged in a 13×13 grid. The phantom was injected with bismuth NPs and iodine at different concentrations and scanned using a Biograph Somatom Edge 128 CT scanner. %under varying exposure settings (80–140 kVp, 215–430 mAs, 1 second rotation). 
Acquisition parameters were systematically varied across tube voltages (80, 100, 120, 140kVp), tube currents (215 and 430mAs), rotation time (1.0s), pitch factors (0.35–1.5), and a fixed slice thickness of 5mm. Images were reconstructed using a 512×512 matrix, while the field of view (FOV), reconstruction kernel, and image orientation were kept constant. Tissue equivalency in CT imaging is frequently assessed by comparing a material’s Hounsfield Unit (HU). Signal-to-noise ratio (SNR) was computed using ImageJ. For each CT image, regions of interest (ROIs) were manually placed over each cylindrical hole and within the main phantom body to represent the background. SNR was calculated as the ratio of the mean intensity to the standard deviation within each ROI ($SNR = \mu/\sigma$). Since both quantities are in Hounsfield Units, the resulting SNR is unitless. In this work, a total of 3,092 images were used, where each image was labelled with image quality metrics. This study focuses exclusively on SNR values. 

The dataset generated and/or analysed during the current study, including images and parameter files, are publicly available at \url{https://github.com/SnehaGeorge22/capri-ct}.

\begin{figure}[t!]
\centering
\includegraphics[width=0.5\textwidth]{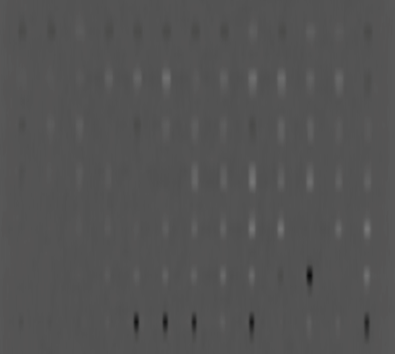}
\caption{Sample CT image taken on perspex phantom} 
\label{fig1}
\end{figure}

%\begin{figure}[htbp]
%    \centering

%    \begin{subfigure}[b]{0.45\textwidth}
%        \centering
        %\includegraphics[width=\textwidth,height=6cm]{phantom.jpg}
        %\caption{The custom phantom used for CT scanning.}
        %\label{fig:phantom}
    %\end{subfigure}
    %\hfill
    %\begin{subfigure}[b]{0.45\textwidth}
     %   \centering
        %\includegraphics[width=\textwidth,height=6cm]{SNR ROI.jpg}
        %\caption{Using imageJ ROI was chosen over the holes.}
        %\label{fig:snr_roi}
    %\end{subfigure}

    %\caption{Phantom and corresponding SNR measurement ROIs.}
    %\label{fig:phantom_combined}
%\end{figure}

\vspace{-10pt}
\section{Methodology}
\vspace{-5pt}

To understand how imaging parameters affect the SNR in CT scans, we introduce a causally aware deep learning approach, which is divided into two main phases: causal discovery and causal learning. 

\subsection{Causal Graph Specification}
In the causal discovery phase, we begin by defining the relevant variables involved in this study. We then discuss our assumptions, which are made in our approach to better understand the underlying causal relationships between imaging parameters and image quality. To formalize these relationships, we employ deep structural equations, which help to model the dependencies among variables. This process culminates in the construction of a causal Directed Acyclic Graph (DAG), which visually represents these cause-effect relationships and guides further analysis.

\textbf{Variables:} The variables observed in this study include tube voltage $v$, tube current $t$, contrast agent types $a$ (such as Iodine, Bismuth Nanoparticles (BiNPs at 100, 50 nanometers in diameter)), acquired CT image $i$, and signal-to-noise ratio $snr$. Furthermore, we consider a latent variable $z$ to capture unmeasured factors that are not explicitly recorded.

\textbf{Assumptions:} Firstly, we assume that a causal context with voltage $v$, tube current $t$, contrast agent type $a$ influences the signal-to-noise ratio. These inputs are modelled as direct causes of the outcome as our proposed CAPRI-CT model learns how changes in these input variables affect the output (SNR), assuming interventions are valid. Secondly, we assume that $z$ is a shared latent representation that influences both the CT image $i$ and consequently the SNR, as both are affected by underlying imaging conditions such as scanner characteristics, phantom specific anatomy, and acquisition noise. Finally, we assume that the latent variable $z$, inferred jointly from the image and acquisition parameters ($v$, $t$, $a$) captures the relevant imaging context. Given $z$ and the explicit parameter embeddings, the raw image provides limited additional information for predicting SNR. This reflects a soft conditional independence, where SNR depends primarily on $z$ and the parameters, rather than the image directly.

\textbf{Deep Structural Equations:} In our proposed CAPRI-CT model, we define the relationships between variables using deep causal structural equations that describe how each variable is associated with each other.

\begin{enumerate}
    \item The CT image $i$ is produced based on a combination of observed imaging parameters such as voltage ($v$), tube current ($t$), and contrast agent type ($a$). The function also includes an exogenous noise term $u_i$ that reflects variability in image acquisition not explained by the other inputs.
    \begin{equation}
    i = f_i(v, t, a, u_i)
    \end{equation}
    
    \item The latent variable $z$ is a shared representation of the image $i$ with observed parameters such as voltage ($v$), tube current ($t$), and contrast agent type ($a$). Here, $u_z$ represents the independent latent noise or exogenous factors that drive the variability in $z$.
    \begin{equation}
    z = f_z(i, v, t, a, u_z)
    \end{equation}
    
    \item The output variable signal-to-noise ratio ($snr$) is calculated as a function of imaging parameters such as voltage ($v$), tube current ($t$), and contrast agent type ($a$) along with the latent variable $z$ and exogenous noise $u_{snr}$.
    \begin{equation}
    snr = f_{snr}(v, t, a, z, u_{snr})
    \end{equation}
\end{enumerate}

The \textbf{Causal Directed Acrylic Graph (DAG)} visually represents how different variables such as voltage, current, and different contrast agents influence the CT image and the final SNR value. The Custom Deep Structural causal model on CT images is shown in Figure \ref{fig1}.

\begin{figure}[t!]
\centering
\includegraphics[width=0.4\textwidth]{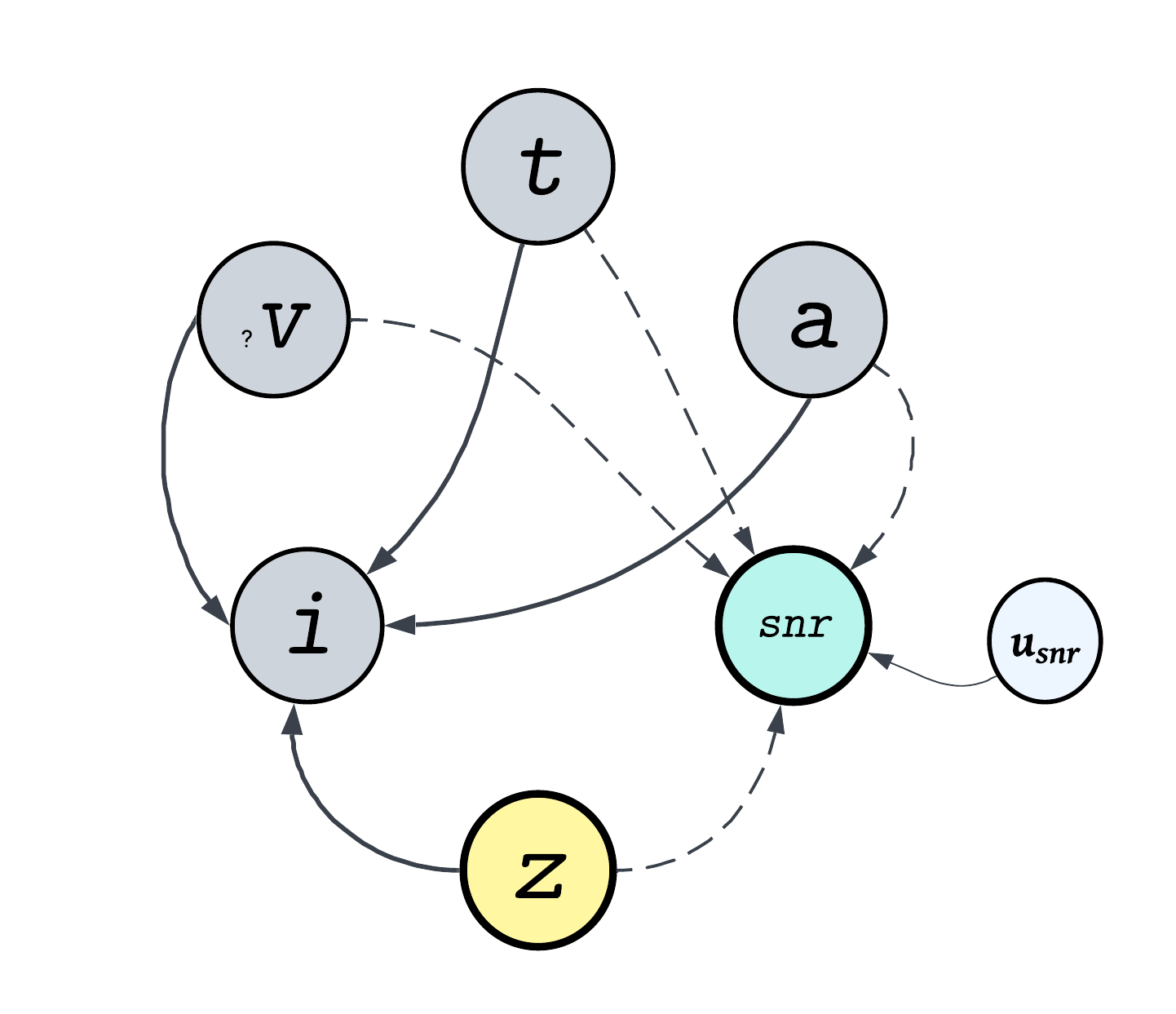}
\caption{This figure represents Causal Directed Acrylic Graph (DAG) of our proposed CAPRI-CT model that captures the causal relationships between signal-to-noise ratio ($snr$) in CT imaging and key observed variables tube voltage ($v$), tube current ($t$), and contrast agent type ($a$) along with latent variable ($z$) and noise ($u_{snr}$). This causal diagram serves as the foundation for both intervention-based and counterfactual reasoning within the CAPRI-CT framework, enabling the estimation of SNR under different scenarios.} \label{fig1}
\end{figure}

\subsection{Causal Learning}

In the causal learning phase of our CAPRI-CT framework, we begin by building a base model using the Variational Autoencoder (VAE) architecture tailored to our specific use case. To enhance robustness, we employ an ensemble approach which includes training and validating multiple instances of the base model. This ensemble allows us to capture the variability in the predictions across models, providing a more reliable estimate of uncertainty. After training, we collect predictions from each model in the ensemble and compute both individual and aggregated performance metrics. The best performing base model is then selected for further analysis. To assess the causal behavior of the model, we conduct interventions and counterfactual inference under different scenarios, and compare the resulting SNR predictions against the original SNR values to evaluate how the model responds to changes in the input variables.

\textbf{Variational Autoencoder Architecture:} The Variational Autoencoder (VAE) is a neural network that learns to compress data into a smaller and meaningful representation and then reconstructs it. It has two main parts: an encoder and a decoder. The \textbf{encoder} is a deep convolutional neural network (CNN) that predicts a distribution by processing the input CT image through three convolutional layers with increasing filter depths (32, 64, 128), each followed by batch normalization and Rectified Linear Units (ReLU) activation. Two stride-2 layers downsample the image spatially, followed by dropout to enhance generalization. The discrete metadata such as voltage, current, and contrast agent type are then embedded into learnable low-dimensional vectors (dimensions 16, 8, and 12 respectively). These embeddings are concatenated and fused with the image features to produce a joint representation. Then, it produces the mean ($\mu$) and log of the variance ($log \sigma$) of that distribution. Then, we implement a \textbf{reparametrization trick} to sample the latent value $z$ from the distribution $N(\mu,diag(\sigma))$ by computing $z=\mu + \sigma . \epsilon$ (element wise product). Finally, the sampled latent variable $z$ is propagated through the regression \textbf{decoder} that is composed of two hidden layers and an output head. This decoder estimates the SNR, enabling both factual prediction and causal inference via interventions or counterfactuals.
\begin{figure}[t!]
\centering
\includegraphics[scale=0.3]{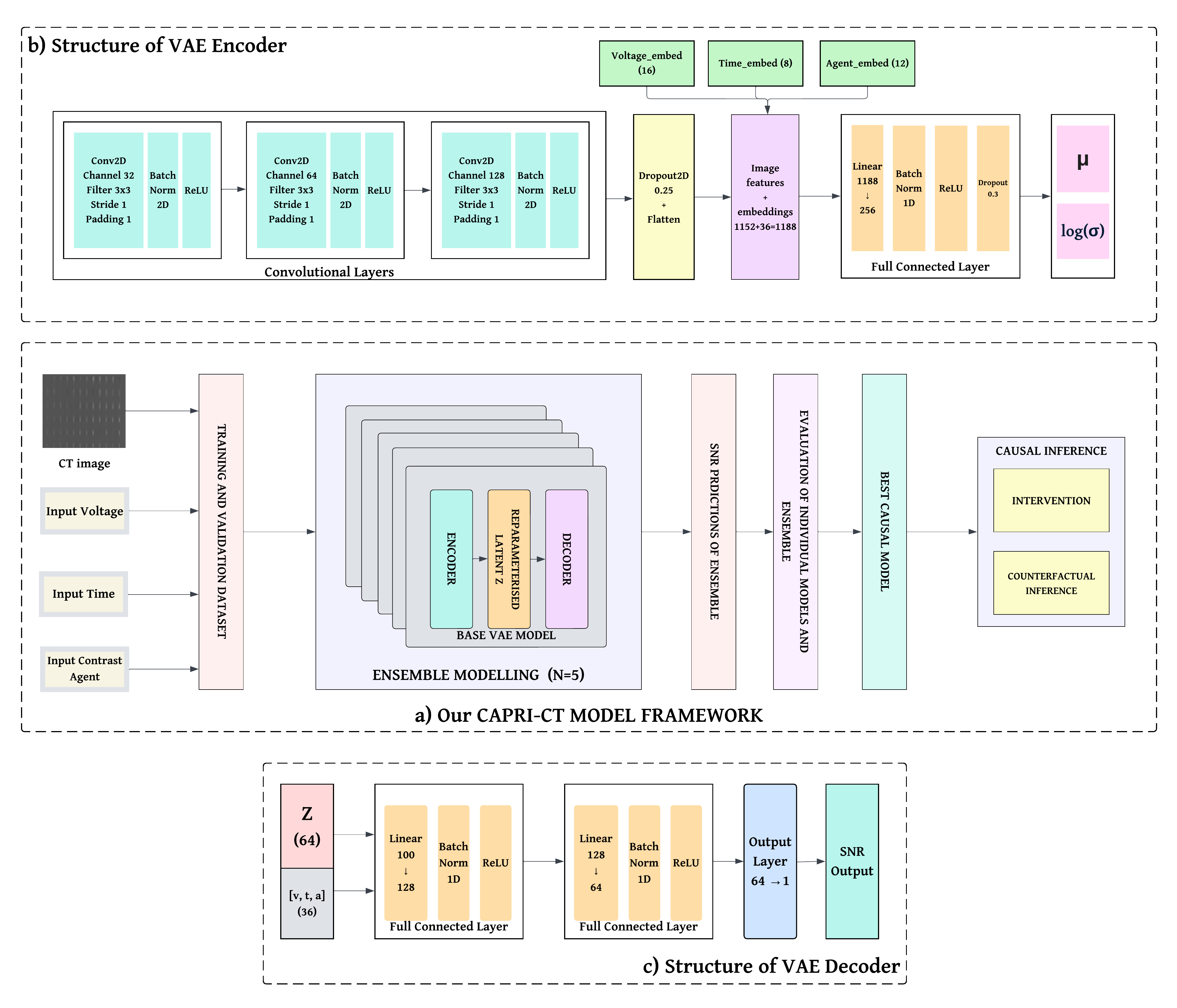}
%\caption{Overview of CAPRI-CT architecture with Ensemble modeling approach for Causal reasoning capabilities a) Architecture of CAPRI-CT framework with Variational Autoencoder(VAE) base model used for the prediction of Signal-to-noise ratio using input CT images and input parameters such as voltage $v$, current $t$, contrast agent $a$ b) Structure of VAE encoder used for extracting CT image features along embedded input variables to create latent variable $z$ c) Structure of VAE decoder that accepts latent variable $z$ along with input variables to predict signal-to-noise ratio.} 
\caption{Overview of the CAPRI-CT architecture with an ensemble modeling approach for causal reasoning. (a) CAPRI-CT uses a VAE-based model to predict signal-to-noise ratio (SNR) from CT images and input parameters: voltage ($v$), current ($t$), and contrast agent ($a$). (b) The VAE encoder extracts latent variable $z$ from CT features and embedded inputs. (c) The decoder predicts SNR using $z$ and the input variables ($v$, $t$, $a$).} \label{fig2}
\end{figure}

\textbf{Ensemble Approach:} To enhance the robustness and reliability of causal inference (and more importantly improve model selection for causal inference), we employ an ensemble-based learning strategy within the CAPRI-CT framework. Multiple instances of the base VAE model are independently trained with varying initial conditions (i.e. random seeds), resulting in a diverse set of trained models. This ensemble of models allows us to capture a broader representation of the underlying causal structure by estimating the prediction uncertainty. Once trained, predictions from all models in the ensemble are aggregated to compute mean outputs and uncertainty bounds (e.g., standard deviation across predictions), providing a measure of confidence in the inferred causal relationships. This ensemble setup not only boosts predictive performance but also enhances the credibility of interventions and counterfactual inferences by accounting for model variance. To evaluate the performance of the model, we use Mean Absolute Error (MAE), Root Mean Squared Error (RMSE) and the coefficient of determination ($R^2$) as performance metrics. Since the target variable (SNR) is unitless, both MAE and RMSE are also reported as unitless quantities. Let $\hat{y}_i$ denote the predicted SNR value and $y_i$ the true SNR value for the $i$-th sample. The MAE loss function measures the average absolute difference between the predicted and ground truth values.
\begin{equation}
\text{MAE} = \frac{1}{n} \sum_{i=1}^{n} |y_i - \hat{y}_i |
\label{eq:mae}
\end{equation}
where $n$ is the total number of samples, $y_i$ is the actual value, $\hat{y_i}$ is the predicted value. The RMSE loss function measures the square root of the average squared difference between the predicted and ground truth values.
\begin{equation}
\text{RMSE} = \sqrt{\frac{1}{n} \sum_{i=1}^{n} (y_i - \hat{y}_i)^2}
\label{eq:mse}
\end{equation}
The $R^2$ or the coefficient of determination metric evaluates the proportion of variance in the SNR that is predictable from the independent variables.

\begin{equation}
R^2 = 1 - \frac{\sum_{i=1}^{n}(y_i - \hat{y}_i)^2}{\sum_{i=1}^{n}(y_i - \bar{y})^2}
\label{eq:r2}
\end{equation}
where $\bar{y}$ is the mean of the true SNR values.

\textbf{Causal Inference:} As a final phase, we begin by selecting the best performing causal base model, which has the highest value of the performance metric, the coefficient of determination ($R^2$), from the ensemble of trained models. We then perform interventions and counterfactual inference using this model to analyse how the input variables affect the output SNR. 
An \textbf{intervention} is defined as the process of actively modifying one or more input variables to observe the resulting effect on the outcome, simulating a controlled experiment (often denoted using the do-operator in causal theory). \textbf{Counterfactual inference} is defined as the estimation of what the outcome would have been under a different set of input conditions, given what actually happened essentially asking ``what if'' questions. By comparing the predicted SNR after an intervention or counterfactual adjustment to the original SNR value, we gain insights into the causal impact of each variable on image quality. Figure \ref{fig2} illustrates the architecture. 

\textbf{Causal Identifiability and Simulation:} Our proposed model aims to estimate how input parameters such as voltage ($v$), current ($t$), and contrast agent ($a$) causally affect image quality, measured by signal-to-noise ratio $snr$. Formally, we want to evaluate interventional distributions of the form:
\begin{equation}
P(snr \mid  do(v),do(t),do(a))
\end{equation}
which represent the causal effect of setting input parameters to specific values. All parents of $snr$ (namely $v$, $t$, $a$, and $i$) are observed, ensuring the identifiability of the causal effect via backdoor adjustment. During training, CAPRI-CT learns the conditional distribution
\begin{equation}
P(snr \mid  v,t,a,i).
\end{equation}
During simulation, for instance, CAPRI-CT simulates interventions such as $do(v=120)$ by overriding the voltage in the model’s input and recomputing the predicted SNR.
\begin{equation}
    \hat{y}_{do(v=120)} = f(i, v'=120, t, a),
\end{equation}
where $v'$ denotes the intervened voltage value. This embedding manipulation directly corresponds to the \textit{truncated factorization formula} from Pearl’s do-calculus:
\begin{equation}
    P(snr \mid do(v=120)) = \sum_{t, a, i} P(snr \mid v=120, t, a, i) \cdot P(t, a, i),
\end{equation}
which allows estimating interventional quantities from observational data without retraining the model.

\section{Experiment Study}

The dataset was split 80/20 into training and validation sets using stratified sampling by SNR quantiles to ensure balanced SNR distribution. Training images underwent data augmentation (rotations, flips), while validation images were only resized. Rare extreme SNR samples were duplicated and assigned higher sampling weights to address imbalance, with a weighted random sampler used during training. Categorical metadata (voltage, current, contrast agent) were encoded numerically.

The training results of the CAPRI-CT ensemble model for the regression task of predicting SNR values from CT phantom images are summarized in Table 1. The ensemble approach achieved strong predictive performance, attaining an $R^{2}$ value of approximately 0.796 and an RMSE of 107.318 on the validation set. Among the five individually trained models that formed the ensemble, Model 5 performed the best, achieving an $R^{2}$ of 0.799, RMSE of 106.493, and MAE of 68.028. Notably, all five models demonstrated consistent performance, with $R^{2}$ values ranging from 0.790 to 0.799. %The ensemble further enhanced prediction stability and reduced error, yielding the lowest MAE of 67.714 across all evaluations. 

Compared to CNN Baseline and ResNet, SqueezeNet showed the most stable performance across five folds, with MAE (91.45–102.93), RMSE (140.08–148.71), and $R^2$ (0.608–0.669). While the proposed CAPRI-CT model achieved higher overall accuracy, SqueezeNet’s low variance highlights its robustness as a reliable lightweight baseline for SNR estimation in CT imaging.
%We compared the proposed CAPRI-CT framework against several baseline models, including CNN Baseline, ResNet and SqueezeNet. Among the baseline models, SqueezeNet demonstrated the most stable performance across all five folds. The model’s MAE values ranged narrowly from 91.45 to 102.93, and RMSE values from 140.08 to 148.71. The $R^2$ scores also remained consistent, varying between 0.608 and 0.669. This tight clustering of error metrics indicates that SqueezeNet produced reliable and reproducible results across different runs. Although its overall accuracy was lower than that of CAPRI-CT, its low variance highlights its robustness as a lightweight baseline model for SNR estimation in CT imaging.

Statistical analysis using the Friedman test confirmed significant differences in model performance (MAE: $\chi^2 = 9.72$, $p = 0.0211$; RMSE: $\chi^2 = 14.04$, $p = 0.0029$). Wilcoxon signed-rank tests showed CAPRI-CT significantly outperformed SqueezeNet in both MAE and RMSE (statistic = 0.0, $p = 0.0312$), demonstrating superior accuracy and consistency.

To better understand the model’s internal representations, correlation analysis between the latent variables and SNR was conducted. The results revealed that contrast agent had a strong influence on SNR, while voltage and current showed minimal correlation; however, since these are known direct causes of SNR, and no external factors such as phantom or scanner details were included in the model, potential confounders could not be identified or confirmed. This is a limitation of the dataset used in this research.

%\vspace{-10pt}
\begin{table}[t!]
\centering
\caption{Comparison of method accuracy}
\begin{tabular}{|l|l|l|l|l|}
\hline
Architecture & Epochs & MAE & RMSE & R\textsuperscript{2} \\
\hline
CAPRI-CT (ours)  & 54  & 68.0280  & 106.4930  & 0.7990 \\
CNN (basline)             & 26  & 94.7015  & 141.2704  & 0.6773 \\
ResNet           & 93  & 89.3858  & 129.1232  & 0.7502 \\
Squeezenet       & 93  & 94.2543  & 139.2030  & 0.6867 \\
\hline
\end{tabular}
\label{tab:model_comparison}
\end{table}
%\vspace{-10pt}

\begin{figure}[t!]
\centering
\includegraphics[width=0.8\textwidth]{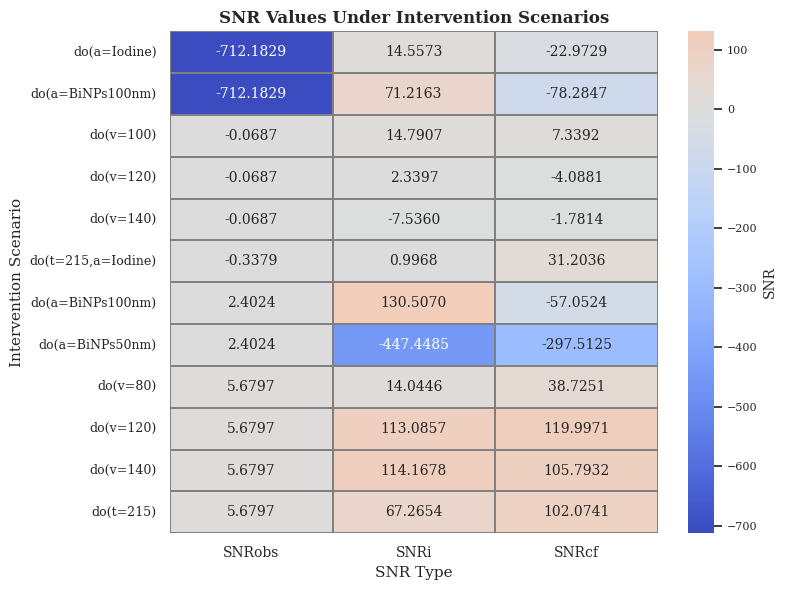}
\caption{Heatmap of SNR estimates - $SNR_{obs}$, $SNR_{i}$, and $SNR_{cf}$} 
\label{fig3}
\end{figure}

To elucidate the causal reasoning capabilities of the CAPRI-CT model, we performed a comprehensive analysis using interventional and counterfactual inference techniques across variations in voltage and contrast agent. As shown in Table~2, the model demonstrates robust causal awareness. For example, substituting the agent from BiNPs 50nm to Iodine at 100 kVp and 215 mAs shifted the observed SNR from $-712.18$ to an intervened value of $14.56$, while the counterfactual SNR remained negative at $-22.97$, reflecting a notable shift due to the simulated intervention. Likewise, increasing voltage from 80 to 120 kVp for Iodine at 215 mAs improved the intervened SNR from $14.79$ to $2.34$, yet the counterfactual SNR declined from $7.34$ to $-4.09$, suggesting complex non-linear interactions. These results affirm that the CAPRI-CT model not only performs accurate regression but also supports causal diagnostics critical for informed protocol optimization in CT imaging. Fig. \ref{fig3} shows SNR values under different intervention scenarios. $SNR_{i}$ and $SNR_{cf}$ deviate significantly from $SNR_{obs}$ for certain interventions (e.g., do(a = BiNPs50nm)), highlighting the model’s sensitivity to contrast agent changes. Voltage-only interventions show moderate effects, indicating the model’s ability to isolate and quantify causal impacts on SNR.
%\vspace{-10pt}
\begin{table}[t!]
\centering
\caption{SNR Evaluation under Interventions and Counterfactual Scenarios}
\begin{tabular}{@{}cccclll@{}}
\toprule
\textbf{v} & \textbf{t} & \textbf{a} & $\boldsymbol{SNR}_{\mathit{obs}}$ & \textbf{What-if scenario} & $\boldsymbol{SNR}_{\mathit{i}}$ & $\boldsymbol{SNR}_{\mathit{cf}}$ \\
\midrule
100 & 215 & BiNPs 50nm   & -712.1829 & do($a=\mathrm{Iodine}$)               & 14.5573   & -22.9729 \\
100 & 215 & BiNPs 50nm   & -712.1829 & do($a=\mathrm{BiNPs\ 100nm}$)         & 71.2163   & -78.2847 \\
80  & 215 & Iodine       & -0.0687   & do($v=100$)                           & 14.7907   & 7.3392   \\
80  & 215 & Iodine       & -0.0687   & do($v=120$)                           & 2.3397    & -4.0881  \\
80  & 215 & Iodine       & -0.0687   & do($v=140$)                           & -7.5360   & -1.7814  \\
120 & 430 & BiNPs 100nm  & -0.3379   & do($t=215$,$a=\mathrm{Iodine}$) & 0.9968    & 31.2036  \\
140 & 430 & Iodine       & 2.4024    & do($a=\mathrm{BiNPs\ 100nm}$)         & 130.5070  & -57.0524 \\
140 & 430 & Iodine       & 2.4024    & do($a=\mathrm{BiNPs\ 50nm}$)          & -447.4485 & -297.5125 \\
100 & 430 & BiNPs 100nm  & 5.6797    & do($v=80$)                            & 14.0446   & 38.7251  \\
100 & 430 & BiNPs 100nm  & 5.6797    & do($v=120$)                           & 113.0857  & 119.9971 \\
100 & 430 & BiNPs 100nm  & 5.6797    & do($v=140$)                           & 114.1678  & 105.7932 \\
100 & 430 & BiNPs 100nm  & 5.6797    & do($t=215$)                           & 67.2654   & 102.0741 \\
\bottomrule
\end{tabular}
\label{tab:snr_intervention}
\end{table}
%\vspace{-10pt}

\begin{table}[ht]
\centering
\caption{Performance of different CAPRI-CT model variants}
\begin{tabular}{lccc}
\toprule
\textbf{Versions of CAPRI-CT model} & \textbf{MAE} & \textbf{RMSE} & \textbf{R\textsuperscript{2}} \\
\midrule
Image (\textit{i}) + metadata (\textit{v, t, a}) & 68.028 & 106.493 & 0.799 \\
Image (\textit{i}) + metadata (\textit{v, t, a, noise}) & 68.704 & 107.192 & 0.797 \\
Image (\textit{i}) + metadata (\textit{v, a})     & 87.209 & 133.561 & 0.684 \\
Image (\textit{i}) + metadata (\textit{t, a})     & 91.421 & 139.720 & 0.655 \\
Image (\textit{i}) only                           & 172.722 & 235.216 & 0.021 \\
Image (\textit{i}) + metadata (\textit{v, t})     & 164.256 & 237.137 & 0.005 \\
\bottomrule
\end{tabular}
\label{tab:capri-ct-ablation}
\end{table}

\textbf{Ablation Study and Causal Perturbation Analysis:} Table \ref{tab:capri-ct-ablation} reports the results of causal structure perturbation through ablation studies, evaluating the impact of removing individual input variables on the model's performance. The full CAPRI-CT model achieves the best predictive accuracy (MAE: 68.03, RMSE: 106.49, $R^2$
: 0.799). Ablating current (t) and voltage (v) led to moderate degradation in performance, with $R^2$ decreasing to 0.684 and 0.655, respectively. This indicates their contribution to the prediction task but suggests limited causal influence relative to other variables. By contrast, removing the contrast agent (a) resulted in a substantial drop in performance ($R^2$: 0.005), nearly equivalent to removing all three inputs ($R^2$: 0.021). To further test model robustness, we introduced an additional noise variable to each input parameter; performance remained stable (MAE: 68.70, RMSE: 107.19, $R^2$: 0.797). This highlights the contrast agent as a dominant causal factor for SNR in CT imaging. These results support the causal assumptions embedded in the model and demonstrate its sensitivity to disruptions in key parent nodes of the causal graph.

\section{Conclusions and Discussion}

This paper demonstrates that optimizing CT imaging parameters (voltage, current, and contrast agent) can significantly improve SNR. Causal analysis revealed that many low-SNR outcomes stem from suboptimal settings, not hardware limits, and highlighted complex parameter interactions best captured through interventional and counterfactual reasoning.

%The causal analysis presented in this study highlights the importance of jointly optimizing imaging parameters specifically voltage, current, and contrast agent choice—to improve signal-to-noise ratio (SNR) in biomedical imaging. Interventional results demonstrate that small changes in these parameters can lead to significant gains in SNR, particularly under specific voltage and current settings. Notably, counterfactual estimates revealed substantial latent performance potential in cases where the observed SNR was low, indicating that many poor outcomes could be attributed to suboptimal parameter combinations rather than intrinsic limitations. The non-linear interactions observed between voltage and agent type further emphasize the value of causal modeling in understanding complex dependencies that traditional correlation-based approaches may overlook. By leveraging interventional and counterfactual reasoning, our approach enables data-driven protocol optimization prior to physical experimentation.

While this study focused specifically on SNR prediction, the proposed CAPRI-CT framework could be extended to other image quality metrics such as contrast-to-noise ratio, sharpness, or artifact levels. Clinically, such models could support decision-making by simulating whether alternative scan settings might improve image quality, potentially helping avoid unnecessary repeat scans. Although this study held certain technical parameters constant such as reconstruction kernels and window width/level (WW/WL) these factors also influence SNR and could be incorporated in future extensions for a more comprehensive analysis. Additionally, while the model identifies potential confounders affecting image quality, it does not explicitly adjust for them during training but rather focuses on simulating 'what-if' scenarios to explore how changes in scan parameters might impact outcomes.

% \vspace{0.1cm}
% \noindent
% For the purpose of open access, the authors have applied a Creative Commons Attribution (CC BY) licence to any Author Accepted Manuscript version arising from this submission. %% don't delete this -- we need it for Exeter publications to be compliant with OA rules! Putting in now so it's not forgotten: JF


\bibliographystyle{unsrt}
\bibliography{references}

\begin{thebibliography}{10}

\bibitem{selfsupervised2019}
L.~Chen, P.~Bentley, K.~Mori, et~al.
\newblock Self-supervised learning for medical image analysis using image context restoration.
\newblock {\em Medical Image Analysis}, 58:101539, 2019.

\bibitem{goldphantom2021}
M.~Oumano, L.~Russell, M.~Salehjahromi, et~al.
\newblock {CT imaging of gold nanoparticles in a human-sized phantom}.
\newblock {\em Journal of Applied Clinical Medical Physics}, 22(1):337--342, 2021.

\bibitem{mambo2025}
J.~Li, S.~Wang, K.~Zhou, et~al.
\newblock Mambo-net: Multi-causal aware modeling backdoor-intervention optimization for medical image segmentation network.
\newblock {\em arXiv preprint}, arXiv:2505.21874, 2025.

\bibitem{causalitymatters2020}
D.C. Castro, I.~Walker, and B.~Glocker.
\newblock Causality matters in medical imaging.
\newblock {\em Nature Communications}, 11:3673, 2020.

\bibitem{blackbox2025}
N.~Sani, D.~Malinsky, and I.~Shpitser.
\newblock Explaining the behavior of black-box prediction algorithms with causal learning.
\newblock {\em arXiv preprint arXiv:2006.02482}, 2020.

\bibitem{review2022}
A.~Vlontzos, D.~Rueckert, and B.~Kainz.
\newblock A review of causality for learning algorithms in medical image analysis.
\newblock {\em Journal of Machine Learning for Biomedical Imaging. pp 1-17}, 2022.

\bibitem{pawlowski2020}
N.~Pawlowski et~al.
\newblock Deep structural causal models for tractable counterfactual inference.
\newblock {\em Advances in neural information processing systems. 857-869}, 33, 2020.

\bibitem{fundus2022}
S.~Liao.
\newblock Counterfactual inference on retina fundus images using deep structural causal models.
\newblock Master's thesis, 2022.

\bibitem{latent3d2024}
W.~Peng, T.~Xia, F.D.S. Ribeiro, et~al.
\newblock {Latent 3D Brain MRI Counterfactual}.
\newblock {\em arXiv preprint}, arXiv:2409.05585, 2024.

\bibitem{staging2023}
S.~Lee, K.H. Lee, J.H. Park, H.Y. Kim, Y.~Choi, and K.H. Lee.
\newblock {Staging chest CT in patients with early-stage colon cancer: analysis of impact on survival using inverse probability weighting and causal diagram}.
\newblock {\em American Journal of Roentgenology}, 221(2):184--195, 2023.

\bibitem{caussl2023}
J.~Miao et~al.
\newblock Caussl: Causality-inspired semi-supervised learning for medical image segmentation.
\newblock In {\em Proceedings of the International Conference on Computer Vision (ICCV)}, 2023.

\bibitem{nofairlunch2023}
C.~Jones, D.C. Castro, F.D.S. Ribeiro, O.~Oktay, M.~McCradden, and B.~Glocker.
\newblock No fair lunch: a causal perspective on dataset bias in machine learning for medical imaging.
\newblock {\em arXiv preprint arXiv:2307.16526}, 2023.

\bibitem{kaddour2022}
J.~Kaddour, A.~Lynch, Q.~Liu, et~al.
\newblock Causal machine learning: A survey and open problems.
\newblock {\em arXiv preprint}, arXiv:2206.15475, 2022.

\bibitem{highstakes2022}
S.~Dash, V.N. Balasubramanian, and A.~Sharma.
\newblock Evaluating and mitigating bias in image classifiers: A causal perspective using counterfactuals.
\newblock In {\em Proceedings of the IEEE/CVF winter conference on applications of computer vision}, pages 915--924, 2022.

\bibitem{highstakes2022b}
B.~Sahoh, K.~Haruehansapong, and M.~Kliangkhlao.
\newblock Causal artificial intelligence for high-stakes decisions: The design and development of a causal machine learning model.
\newblock {\em IEEE Access}, 10:24327--24339, 2022.

\bibitem{multiatt2024}
S.~Huang, L.~Wang, J.~Liao, and L.~Liu.
\newblock Multi-attentional causal intervention networks for medical image diagnosis.
\newblock {\em Knowledge-Based Systems}, 299:111993, 2024.

\bibitem{deepcausalpancreas2024}
C.~Li, Y.~Mao, S.~Liang, J.~Li, Y.~Wang, and Y.~Guo.
\newblock {Deep causal learning for pancreatic cancer segmentation in CT sequences}.
\newblock {\em Neural Networks}, 175:106294, 2024.

\end{thebibliography}
\end{document}